\documentclass[10pt,twocolumn,letterpaper]{article}

\usepackage{iccv}              % To produce the CAMERA-READY version

\makeatletter
\@namedef{ver@everyshi.sty}{}
\makeatother
\usepackage{tikz}
\usepackage{placeins}
\usepackage{multirow}
\definecolor{darkgreen}{rgb}{0.0, 0.5, 0.0}

\usepackage{amsmath}
\usepackage{pifont}
\usepackage[utf8]{inputenc}
\definecolor{iccvblue}{rgb}{0.21,0.49,0.74}
\def\confName{ICCV}
\def\confYear{2025}

\title{ViCTr: Vital Consistency Transfer for Pathology Aware Image Synthesis}

\author{
Onkar Susladkar\textsuperscript{1,3},
Gayatri Deshmukh\textsuperscript{1},
Yalcin Tur\textsuperscript{2},
Gorkem Durak\textsuperscript{1},
Ulas Bagci\textsuperscript{1} \\
\textsuperscript{1}Northwestern University \quad
\textsuperscript{2}Stanford University \quad \textsuperscript{3}University of Illinois Ubrana-Champaign \\
{\tt\small onkarsus13@gmail.com, dgayatri9850@gmail.com, yalcintr@stanford.edu,} \\
{\tt\small gorkem.durak@northwestern.edu, ulas.bagci@northwestern.edu}
}

\begin{document}
\maketitle
\begin{abstract} 
% \vspace{-1.5em}

     Synthesizing medical images remains challenging due to limited annotated pathological data, modality domain gaps, and the complexity of representing diffuse pathologies such as liver cirrhosis. Existing methods often struggle to maintain anatomical fidelity while accurately modeling pathological features, frequently relying on priors derived from natural images or inefficient multi-step sampling. In this work, we introduce \textbf{ViCTr} (\textbf{Vi}tal \textbf{C}onsistency \textbf{Tr}ansfer), a novel two-stage framework that combines a rectified flow trajectory with a Tweedie-corrected diffusion process to achieve high-fidelity, pathology-aware image synthesis. First, we pretrain \textbf{ViCTr} on the ATLAS-8k dataset using \textbf{Elastic Weight Consolidation (EWC)} to preserve critical anatomical structures. We then fine-tune the model adversarially with \textbf{Low-Rank Adaptation (LoRA)} modules for precise control over pathology severity. By \textbf{reformulating Tweedie’s formula} within a linear trajectory framework, \textbf{ViCTr} supports one-step sampling—reducing inference from 50 steps to just 4—without sacrificing anatomical realism. We evaluate \textbf{ViCTr} on BTCV (CT), AMOS (MRI), and CirrMRI600+ (cirrhosis) datasets. Results demonstrate state-of-the-art performance, achieving a Medical Fréchet Inception Distance (MFID) of 17.01 for cirrhosis synthesis—28\% lower than existing approaches—and improving nnUNet segmentation by +3.8\% mDSC when used for data augmentation. Radiologist reviews indicate that \textbf{ViCTr}-generated liver cirrhosis MRIs are clinically indistinguishable from real scans. To our knowledge, \textbf{ViCTr} is the first method to provide fine-grained, pathology-aware MRI synthesis with graded severity control, closing a critical gap in AI-driven medical imaging research.
  
\end{abstract}
    
\section{Introduction}
\label{sec:introduction_Section}
\begin{figure*}[t]
    \centering
    % \hspace{-1cm}
    \includegraphics[width=1.0\textwidth]{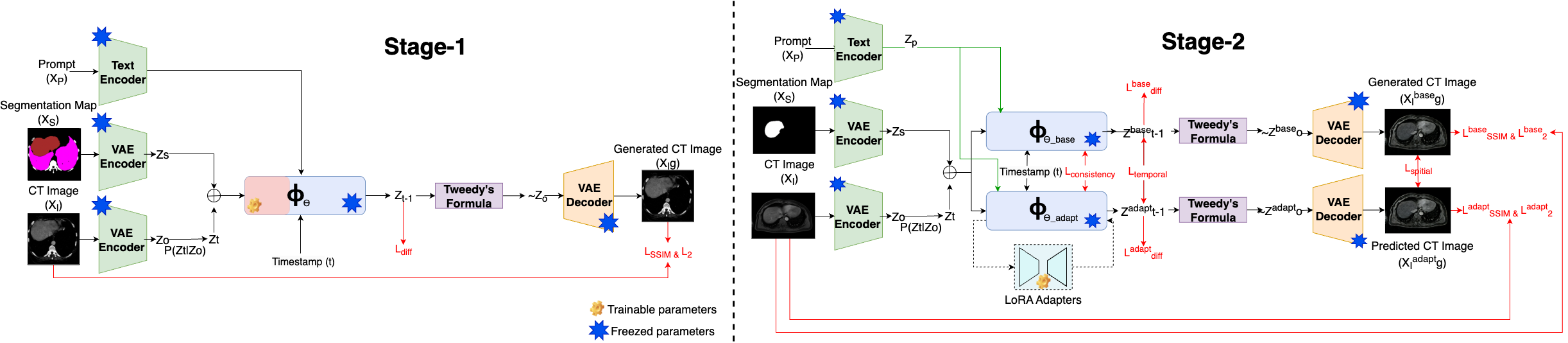}    
    \caption{Overview of the proposed \textbf{ViCTr} methodology.}
    \label{fig:ViCTor}
\end{figure*}
 
The exponential growth in computer vision capabilities has been driven by significant advances in artificial intelligence models~\cite{CHAI2021100134,9903420}. However, medical imaging faces a fundamental tension between model complexity and data availability that limits the application of state-of-the-art techniques. While recent breakthroughs in generative AI have demonstrated remarkable capabilities in synthetic data creation, these advances demand training datasets of unprecedented scale—a requirement that poses unique challenges in the medical domain~\cite{Mahmood_2022_CVPR, althnian2021impact}.

Unlike general computer vision applications, medical imaging faces several critical constraints: privacy regulations necessitating complex deidentification, inherent data fragmentation across healthcare institutions, and fundamental interoperability constraints. These barriers have created a growing disparity between the rapid advancement of general computer vision and the relatively slower progress in medical imaging applications. Current approaches to bridging this gap face two major limitations: insufficient feature preservation across anatomical structures and inadequate handling of pathological variations.

This challenge is especially pronounced in \textbf{abdominal imaging}, where pathologies such as diffuse cirrhosis or multi-tissue disease processes manifest across multiple organ systems. Unlike the well-defined boundaries typical of tumors or simpler structures like bones in X-ray imaging, abdominal pathologies often involve \textbf{subtle and heterogeneous changes} in tissue characteristics, requiring more nuanced feature extraction and synthesis. The inherent complexity of MRI signals—spanning multiple sequences and high spatial resolution—presents additional computational hurdles, further distancing it from the more standardized nature of CT or simpler 2D radiographs.

To address these gaps, our work introduces \textbf{ViCTr} (Vital Consistency Transfer), a two-stage framework that facilitates pathology-aware medical image synthesis with strong anatomical fidelity. We specifically target abdominal CT and MRI data, aiming to generate clinically relevant synthetic datasets that capture both normal anatomy and intricate pathological details. By providing robust augmentation material, our approach holds the potential to improve downstream tasks like segmentation and diagnosis, while alleviating issues of data scarcity and privacy constraints in medical imaging. Our contributions are: 

\begin{itemize}\itemsep0pt
        \item \textbf{Novel Two-Stage Framework.} We propose \textbf{ViCTr}, a method that fuses anatomical consistency and pathological realism for CT and MRI synthesis with high diversity.
        \item \textbf{Tweedie’s Formula in Rectified Flow.} We introduce a rectified flow trajectory reformulation of Tweedie’s formula, ensuring accurate initialization and reducing sampling bias.
        % \item \textbf{ViCTr} is based on a novel formulation of Tweedie's formula within a rectified flow trajectory framework.
        \item \textbf{Wide Applicability.} \textbf{ViCTr} integrates seamlessly with various diffusion models, supporting a broad range of medical imaging tasks.
        \item \textbf{One-Step Sampling \& Tweedie’s Corrections.} We reduce computational overhead by cutting the number of diffusion steps from dozens to a handful, expediting inference without sacrificing quality.
        \item \textbf{Quantitative Improvements in Segmentation.} We demonstrate that adding \textbf{ViCTr}-synthesized CT/MRI images to real-world training sets significantly boosts segmentation performance, underscoring the practical benefits of pathology-focused data augmentation.
        \item \textbf{Enhanced Image Fidelity. }\textbf{ViCTr} achieves consistently lower FID (Fréchet Inception Distance) scores across multiple datasets, indicating superior structural and textural coherence.
        \item \textbf{First Abdominal MRI Pathology Synthesis. }To our knowledge, we are the first to generate abdominal MRI pathologies with progressive severity, offering new possibilities for research and clinical applications in medical imaging. 
    \end{itemize}

By emphasizing both anatomical fidelity and pathological variety, \textbf{ViCTr }presents a powerful step toward bridging the gap between limited medical imaging data and the demands of high-performing AI models.

\section{Related Works}
\label{sec:related_work}
\subsection{Generative Models for Medical Data Augmentation}
Synthetic images have long been explored for data augmentation in medical imaging, with Generative Adversarial Networks (GANs) \cite{nie2017medical, kim2021synthesis, NIPS2017_076023ed} initially offering promising results in tasks such as lesion synthesis and modality translation. However, GAN-based methods often struggle with mode collapse, requiring extensive architectural and training refinements to ensure sufficient diversity and realism in the synthesized images \cite{dhariwal2021diffusion, Muller-Franzes2023-zi}.

\subsection{Diffusion Models in Medical Imaging}
Diffusion-based approaches have emerged as a powerful alternative to GANs for high-quality synthetic image generation \cite{trabucco2024effective, bansal2023leaving, boutros2023idiff, dunlap2023diversify}. Their inherent noise-to-image paradigm provides a more stable training regime and can produce richer data variations. In medical imaging, diffusion models \cite{deshmukh2025meddelinea} have been leveraged to improve segmentation and classification performance by generating diverse training samples \cite{Saragih2024, Chen_2024_CVPR, Sharma_2024_CVPR, PANI2024109273, DiffLungs}.
Recent specialized frameworks, including segmentation-guided diffusion \cite{konz2024anatomically} and text-driven generation built upon RadImageNet \cite{zhang2023emit, Mei2022-ru}, highlight the adaptability of diffusion models to specific clinical scenarios. Efforts such as MixUp-enhanced augmentation \cite{lee2024genmix, carratino2022mixup} and domain adaptation methods \cite{meddiffusion} demonstrate that diffusion techniques can mitigate biases and improve generalization in tasks like classification and risk prediction.

\subsection{General-Purpose Diffusion for Data Augmentation}
Various general-purpose diffusion strategies have been applied to further enrich training datasets. DiffuseMix \cite{islam2024diffusemix} employs conditional prompts to blend real and synthetic data while preserving labels, whereas DreamDA \cite{fu2024dreamda} combines diffusion-based perturbations with pseudo-labeling for semantic consistency. DetDiffusion \cite{wang2024detdiffusion} incorporates object-detection attributes into the diffusion process, while Effective Data Augmentation with Diffusion Models \cite{trabucco2023effective} explores diffusion-based techniques to boost performance in few-shot settings. These methods generally aim to enhance data diversity and improve the training of downstream models.

\subsection{Rectified Flow Models}
\begin{figure}[]
        \centering
        \includegraphics[width=\columnwidth]{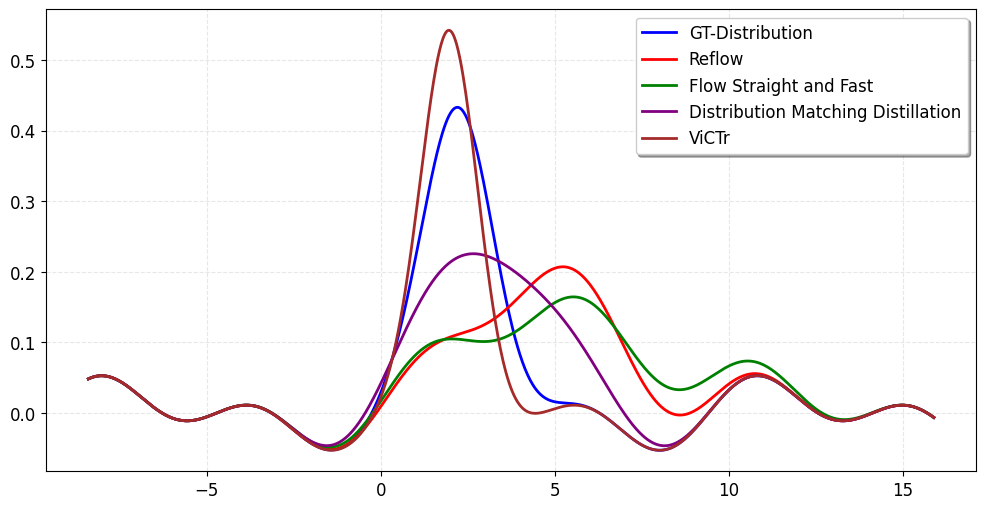}
        \caption{Comparison of synthetic data distributions from various methods against the ground-truth (GT) on the AMOS dataset. The GT distribution (blue) reflects real data, while Reflow (red), Flow Straight and Fast (green), and Distribution Matching Distillation (purple) show varying alignment. Our proposed method, ViCTr (brown) achieves the closest match to GT, demonstrating superior distribution alignment and fidelity.}
        \label{fig:conv}            
\end{figure}

While traditional diffusion approaches rely on discretized noise schedules, Rectified Flow (ReFlow) models learn smooth, near-linear trajectories to map between data distributions. This can result in more efficient sampling and reduced computational overhead. Recent works have leveraged rectified flows for improved convergence rates and lower numbers of function evaluations (NFE) \cite{lee2025improving, liu2022flow, yin2024one}, although most focus on general distribution transport rather than the strict alignment and pathology-aware realism needed in medical imaging. As shown in Figure \ref{fig:conv}, different ReFlow-based methods exhibit varying degrees of distribution matching, with our approach (\textbf{ViCTr}) offering closer alignment to the ground-truth distribution.

\subsection{Domain-Specific Adaptation}
Domain-focused adaptations continue to evolve. DiNO-Diffusion \cite{jimenez2024dino} introduces a self-supervised latent diffusion model to cope with limited annotated medical data, and Diverse Data Augmentation with Diffusions \cite{feng2023diverse} enriches domain generalization using Stable Diffusion combined with cosine similarity filtering. Collectively, these methods illustrate the growing interest in customized diffusion pipelines for specialized domains, especially when high-quality annotated data are scarce.

Recent advancements have further showcased the potential of diffusion models for domain-specific adaptation. DiNO-Diffusion \cite{jimenez2024dino} addresses the challenge of limited annotated data in medical imaging by leveraging a self-supervised latent diffusion framework. Similarly, Diverse Data Augmentation with Diffusions \cite{feng2023diverse} enhances domain generalization by integrating Stable Diffusion with cosine similarity-based filtering, enabling the generation of semantically diverse and high-quality data. However, these methods prioritize general-domain efficiency over the anatomical-pathological alignment critical for medical imaging, often failing to preserve fine-grained structures like vascular networks or diffuse fibrosis patterns.

In this work, we address the critical need for efficient, high-fidelity pathology-aware synthesis in abdominal imaging by merging rectified flow concepts with a Tweedie-corrected diffusion process. Our approach (\textbf{ViCTr}) stands out from prior methods by introducing a linearized sampling framework that enables one-step generation while maintaining anatomical fidelity—a crucial requirement for robust data augmentation in clinical applications.

\section{Methods}

Our framework, \textbf{ViCTr}, uses \textbf{Rectified Flow}~\cite{liu2022flowstraightfastlearning} and \textbf{Tweedie’s Formula}~\cite{Robbins1992} for high-fidelity medical image synthesis with efficient one-step sampling. By rectifying the trajectory from a prior distribution $p_0$ to the target medical image distribution $p_{\text{target}}$, we reduce sampling bias and retain key anatomical and pathological details.

\subsection{Rectified Flow Trajectory}
Classical diffusion approaches gradually corrupt data samples $x \in \mathbb{R}^d$ into a Gaussian prior $p_0$. While these methods have shown success, they often suffer from suboptimal sampling paths, cascading prediction errors, and diminished diversity in generated samples. We address these issues through \textit{Rectified Flow optimization}~\cite{liu2022flowstraightfastlearning, fei2024fluxplaysmusic}, which learns a continuous velocity field guiding the forward and reverse diffusion processes more directly.

Let $(x_0, x_1)$ be a sample pair with $x_0 \sim p_0$ and $x_1 \sim p_{\text{target}}$. We define an interpolated point  $x_t = (1 - t)x_0 + tx_1$ where $t \in [0, 1]$. A velocity model $v_{\theta}: \mathbb{R}^{d} \times [0,1] \rightarrow \mathbb{R}^d$ then predicts how to transition from $x_t$ toward $x_1$. We train $v_{\theta}$ by minimizing:

\begin{equation}
  \hat{\theta} = \arg \min_{\theta} \mathbb{E}_{t \sim \text{Uniform}(0,1)} \bigg[\left\|(x_1 - x_0) - v_{\theta}(x_t, t)\right\|^2 \bigg],
\end{equation}
ensuring the predicted flow $v_{\theta}(x_t, t)$ aligns with the true path $(x_1-x_o)$. Once trained, the rectified flow is realized  by solving the ODE:
\begin{equation}
dx_t = v_{\hat{\theta}}(x_t, t)dt,
\end{equation}
leading $x_0 \sim p_0$ toward $x_1 \sim p_{\text{target}}$. To enable one-step sampling, we distill this multi-step process into a neural network $\hat{\mathcal{N}}: \mathbb{R}^d \rightarrow \mathbb{R}^d$, such that $\hat{\mathcal{T}}(x_0) = x_0 + v(x_0, 0)$, is trained to directly predict $x_1$ from $x_0$. The loss function, 
\begin{equation}
\mathcal{L} = \mathbb{E}\bigg[\left\|(x_1 - x_0) - v(x_0,0)\right\|^2 \bigg],
\end{equation}
drives $\hat{\mathcal{T}(x_0)}$ to approximate the final, rectified state $x_1$.

\subsection{Rectified Flow with Tweedie's Formula}

Though rectified flow refines sampling, medical data distributions often need extra bias correction. Tweedie’s Formula \cite{Robbins1992} addresses this by adjusting noisy observations to better approximate the posterior mean. For Gaussian variables $z \sim \mathcal{N}(\mu_z, \Sigma_z)$, Tweedie's formula indicates: 
$$ \mathbb{E}[\mu_z \mid z] = z + \Sigma_z \nabla_z \log p(z)$$,
where $\nabla_z \log p(z)$ is the gradient of the log probability. In typical diffusion, Tweedie’s formula estimates $x_0$ from noisy $x_t$, guiding predictions toward the data manifold.

\textbf{ViCTr} incorporates this correction into the rectified flow ODE:
$$dx_t = v_{\hat{\theta}}(x_t, t)dt + (1 - \bar{\alpha}_t)\nabla_{x_t} \log p(x_t)dt,$$ and updates the training objective to:
\begin{align*}
\hat{\theta} = \arg \min_{\theta} \, \mathbb{E}_{t \sim \text{Uniform}(0,1)} \bigg[ 
& \big\| (x_1 - x_0) - v_{\theta}(x_t, t) \notag \\
& - (1 - \bar{\alpha}_t) \nabla_{x_t} \log p(x_t) \big\|^2 \bigg].
\end{align*}
Here, $\bar{\alpha}_t$ denotes the cumulative product of variance decay factors in the diffusion schedule. The additional Tweedie term $(1 - \bar{\alpha}_t) \nabla_{x_t} \log p(x_t)$ corrects for sampling bias, driving $x_t$ more accurately toward the target distribution. 

We similarly extend the one-step distillation to integrate Tweedie's correction:
$$\hat{\mathcal{T}}(x_0) = x_0 + v(x_0, 0) + (1 - \bar{\alpha}_0) \nabla_{x_0} \log p(x_0),$$ and optimize $$\mathcal{L} = \mathbb{E}\bigg[\big\| x_1 - \hat{\mathcal{T}}(x_0) \big\|^2 \bigg].$$ This single-step approach balances computational efficiency with high-quality, pathology-aware generation—critical in medical applications where datasets are limited and synthetic realism is paramount.

In summary, \textbf{ViCTr} unifies rectified flow and Tweedie’s correction to deliver anatomically consistent, pathology-aware sampling in a single forward pass. The resulting framework exhibits reduced inference costs, minimized sampling bias, and improved fidelity for generating high-resolution medical images.

\subsection{Pathology Aware Image Synthesis}
\label{sec:proposed_methodology_section}
 
Our proposed \textbf{ViCTr} method adopts a two-stage training paradigm: \textbf{Stage 1} establishes a foundational diffusion model tailored to medical imaging data, and \textbf{Stage 2} fine-tunes this pre-trained model for downstream tasks, including semantic-guided multi-modal generation and counterfactual pathology synthesis.

\subsubsection*{Pre-training on ATLAS-8k Dataset (Stage 1)} \label{stage_1_architecture}
In the absence of large-scale, domain-specific pre-trained diffusion models, we begin by training on the ATLAS-8k \cite{qu2024abdomenatlas} dataset, which comprises abdominal CT scans and their segmentation annotations. We leverage these annotations as conditional guidance—enabling precise control over anatomical structures—and integrate textual prompts (e.g., “create image having 
$<<$organs$>>$”) to further refine semantic consistency.

\textbf{Latent Representations.} Following Figure \ref{fig:ViCTor}, we feed raw CT images $(X_I)$ and segmentation masks $(X_S)$ into a frozen VAE encoder, producing latent embeddings $(Z_o$ and $Z_s$).  Simultaneously, textual prompts $(X_p)$ are processed by a pre-trained text encoder (details in Table \ref{tab:mapping}) to yield prompt embeddings $(Z_p)$. These latent representations guide the generative diffusion backbone $\phi_{\theta}$ during both forward and reverse diffusion.

\textbf{Forward Diffusion.} We progressively inject noise into $Z_o$ forming a noisy representation $Z_t$ as $$P(Z_t | Z_o) = (1 - t) \cdot Z_o + t \cdot \epsilon_{\text{true}},$$ where $\epsilon_{true}$ is the true noise at step $t$. We concatenate $Z_t$ with $Z_s$ and feed them into the diffusion model $\phi_{\theta}$. Crucial layers are selectively unfrozen based on elastic weight consolidation, maintaining model stability while adapting to medical domain specifics.

\textbf{Reverse Diffusion.} In the reverse process, 
   \begin{equation}
        P(Z_{t-1} | Z_t, Z_s, Z_p, t) = Z_t + \delta T \times \phi_{\theta}(Z_t, Z_s, Z_p, t),
    \end{equation} 
iteratively removes noise to reconstruct $Z_{t-1}$ conditioned on the segmentation map $(Z_s)$ and text prompt $(Z_s)$. The diffusion loss $L_{diff}$ between $Z_{t-1}$ and $\epsilon_{true}$ is:
    \begin{equation}
        L_{\text{diff}} = - \left| \phi_{\theta}(Z_t, Z_s, Z_p, t) - (\epsilon_{\text{true}} - Z_o )\right|^2.
    \end{equation}
After refining $Z_{t-1}$ via Tweedie’s formula, the denoised latent $~Z_o$ is fed into the frozen VAE decoder (matching $\phi_{\theta}$; see Table \ref{tab:mapping}) to obtain a reconstructed CT image.

\textbf{Composite Loss}. To ensure both local accuracy and global perceptual realism, we combine: Diffusion Loss ($L_{\text{diff}}$), pixel-level reconstruction loss ($L_2$) evaluating the gap between generated $X_{ig}$ and ground-truth $X_I$, and structural similarity (SSIM) loss ($L_{SSIM}$) emphasizing higher-order textural correspondence. Optimizing this composite loss enables the diffusion backbone $\phi_{\theta}$ to capture essential anatomical features, establishing a versatile medical foundation model ready for fine-tuning.

    \begin{table}[b]
        \centering
        \scriptsize
        \begin{tabular}{ccc}
        \hline
        \textbf{Diffusion Methods} & \textbf{Denoiser}      & \textbf{Text Encoder}           \\ \hline
        Stable Diffusion \cite{Stable-Diffusion-1.5}              & UNet \cite{ronneberger2015u}                   & Clip-B/16  \cite{radford2021learning}                     \\
        Pixart-alpha \cite{pixart-sigma}           & Transformer (DiT) \cite{peebles2023scalable}     & T5-XXXL \cite{raffel2020exploring}                        \\
        Stable Diffusion XL \cite{SD-XL}            & Dual Unet \cite{SD-XL}             & Clip-L/14                       \\
        Flux \cite{fluxdev}             & MultiModal Transformer \cite{Stable-Diffusion-3}  & T5-XXXL + Clip-L/15 \\
        Stable Diffusion-3 \cite{Stable-Diffusion-3}             & MultiModal Transformer   & \begin{tabular}[c]{@{}c@{}}T5-XXXL \ + Clip-B/16 \\ + Clip-L/14 \end{tabular}            \\ \hline
        \end{tabular}
         \caption{Selection of denoisers and text encoders for different diffusion methods $\phi_{\theta}$}
        \label{tab:mapping}

    \end{table}

\subsubsection*{Fine-tuning on downstream tasks (Stage 2)} \label{stage_2_architecture}

Building on the Stage1 pre-trained model, Stage2 targets specialized tasks like semantic-guided CT/MRI generation and counterfactual pathology synthesis. As shown in Figure \ref{fig:ViCTor}, a dual-network setup is adopted.
\begin{enumerate}
    \item $\phi_{base}$ remains frozen, retaining the robust anatomical knowledge acquired in Stage 1.
    \item $\phi_{adapt}$ is augmented with LoRA \cite{hu2021lora} modules within the previously fine-tuned layers. These modules are the only trainable parameters, ensuring targeted and stable adaptation.
\end{enumerate}

\textbf{Training Input \& Consistency Loss.} Similar to Stage 1, each training sample includes prompts, segmentation maps, and either CT or MRI images (depending on the task). A frozen VAE encoder extracts latent representations $Z_s, Z_o$, and $Z_p$. Following forward diffusion of $Z_o$ into noisy $Z_t$, both $\phi_{base}$ and $\phi_{adapt}$ process $[Z_t, Z_s]$ with text embedding $Z_p$. We introduce a consistency loss $L_{consistency}$ to align intermediate outputs of $\phi_{base}$ and $\phi_{adapt}$, ensuring stable adaptation without drifting from core representations.

\textbf{Loss Components.}  Diffusion loss ($L^{\text{base}}_{diff}$, $L^{\text{adapt}}_{diff}$) encourages accurate noise prediction. Temporal consistency loss $L_{temporal}$ preserves smooth transitions across time steps in the reverse diffusion. Spatial consistency loss $L_{spatial}$ ensures alignment in the output reconstructions $(Z^{base}_{t-1}, (Z^{adapt}_{t-1})$. Pixel-level losses ($L_2$ and $L_{SSIM}$) compare generated results to ground-truth scans, enforcing both local fidelity and structural coherence.

    \textbf{CT and MRI Generation.} We adapt to BTCV \cite{fang2020multi} (CT) and AMOS \cite{ji2022amos} (MRI) datasets, using organ segmentation masks for prompts. A standard 80/10/10 partition ensures robust evaluation.

    \textbf{Pathology Generation.} For liver cirrhosis, we employ CirrMRI600+ \cite{jha2024cirrmri600+} dataset (T1/T2 MRIs) with segmentation masks marking liver regions. Prompts specify “low,” “mild,” or “severe” cirrhosis intensity. As in the CT/MRI setting, the training/validation/testing split is 80/10/10. The inference details and inference diagram are in the supplementary material.

This selective LoRA-based fine-tuning allows \textbf{ViCTr} to handle complex tasks—ranging from normal organ generation to pathology synthesis—while preserving the anatomical and text-driven guidance learned in Stage 1. The experiments (Section 4) demonstrate that this two-stage framework consistently yields high-quality, clinically realistic images in both CT and MRI settings, including nuanced liver cirrhosis modeling.

\section{Experiments and Results}
\label{sec:experimental_Section}

    The implementation details for training and sampling are provided in the supplementary material. 
\subsection{Quantitative Results}
\label{subsec:quantitative_results_Section}
\begin{figure}
    \includegraphics[width=\linewidth]{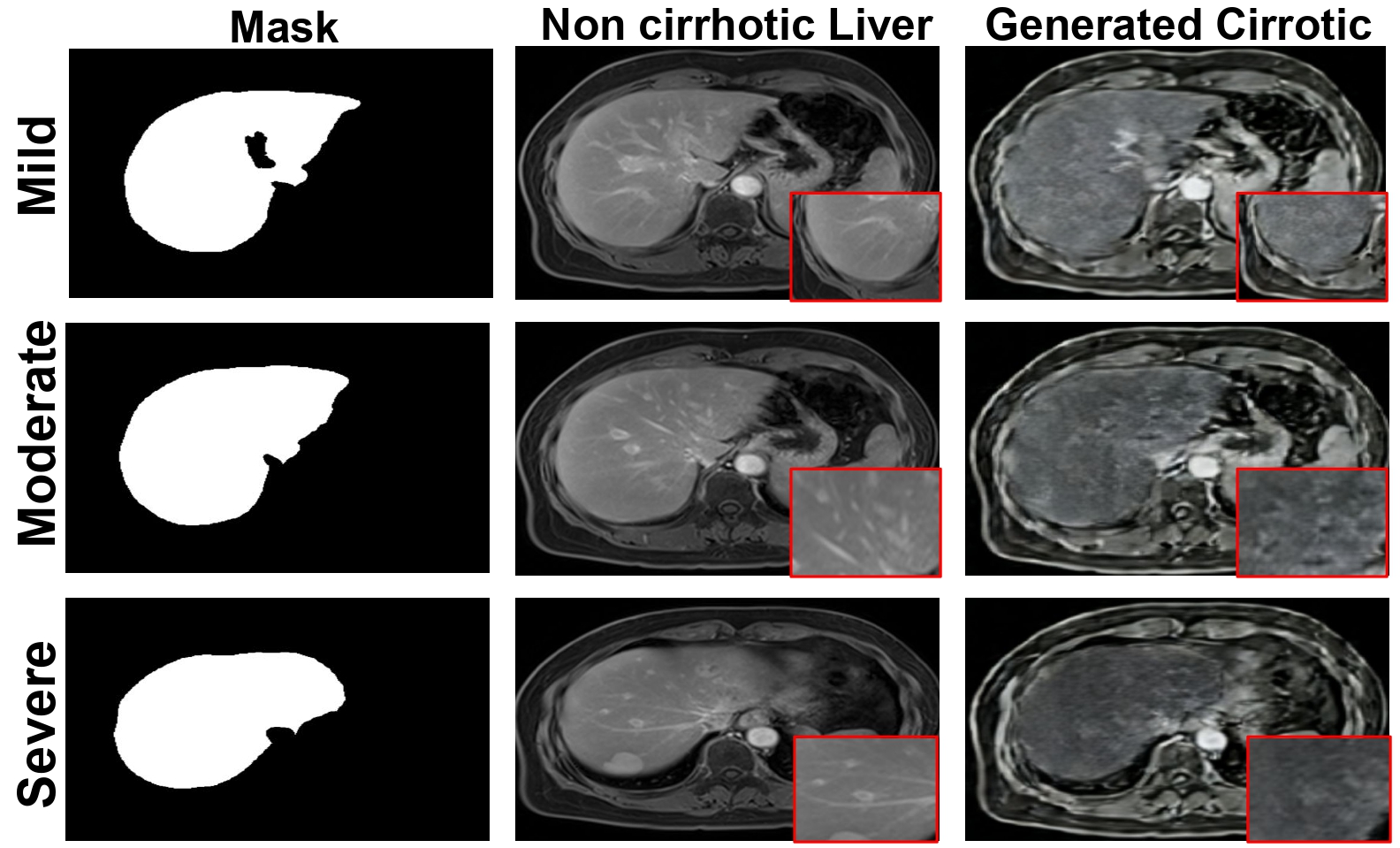}
        \caption{Examples of synthetic cirrhotic images with different severity levels generated from non-cirrhotic scans using \textbf{ViCTr}. For T1-MRI, we show the segmentation mask, the original non-cirrhotic image, and the corresponding synthetic cirrhotic image.}
        \label{fig:cirrosis}

      \includegraphics[width=\linewidth]{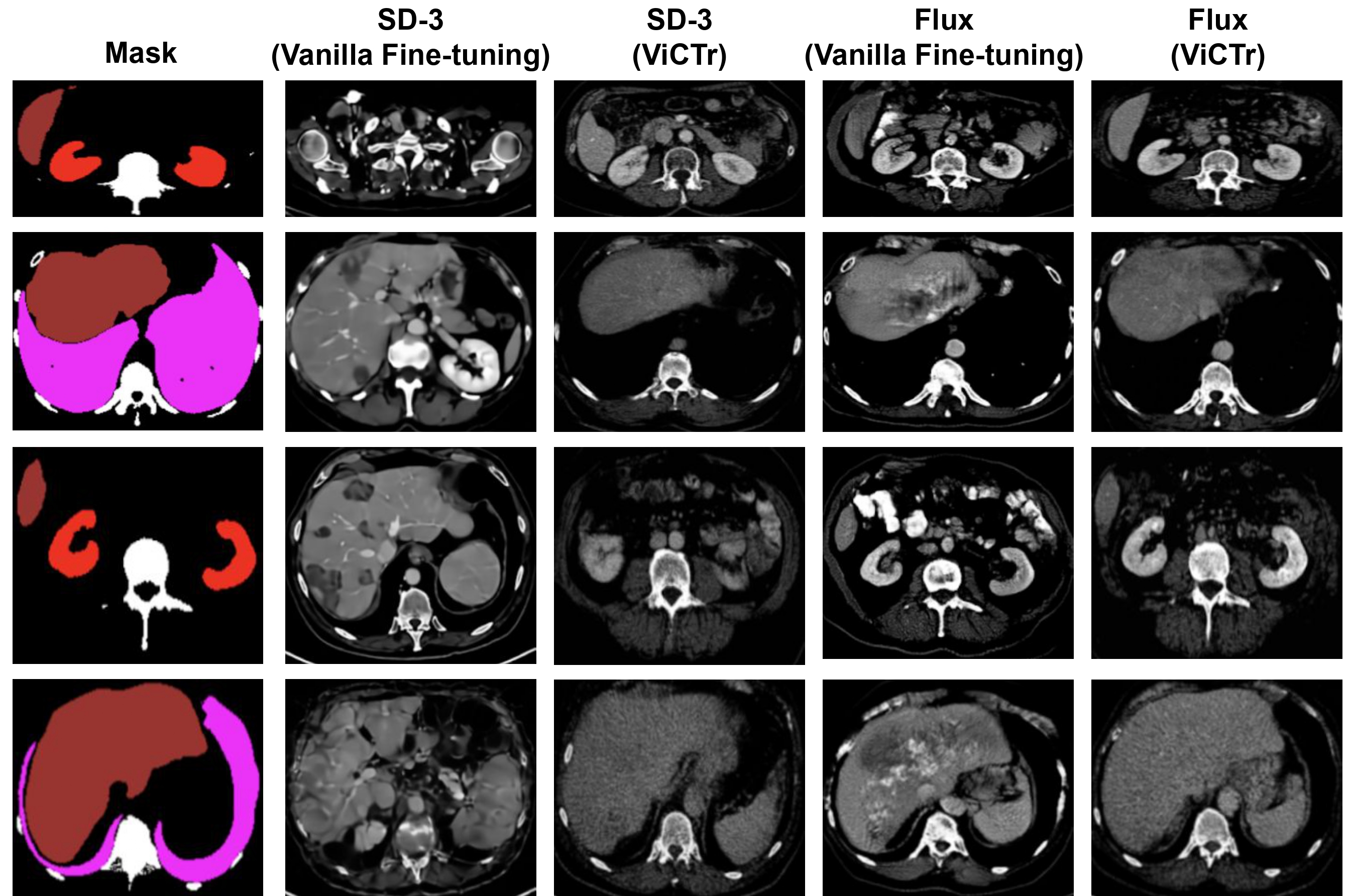}
       \caption{Diffusion-Based Image Generation on BTCV MRI Dataset. From left to right: segmentation mask, SD-3 with standard fine-tuning, FLUX with standard fine-tuning, SD-3 with \textbf{ViCTr}, FLUX with \textbf{ViCTr}.}
        \label{fig:BTCV}

    \includegraphics[width=\linewidth]{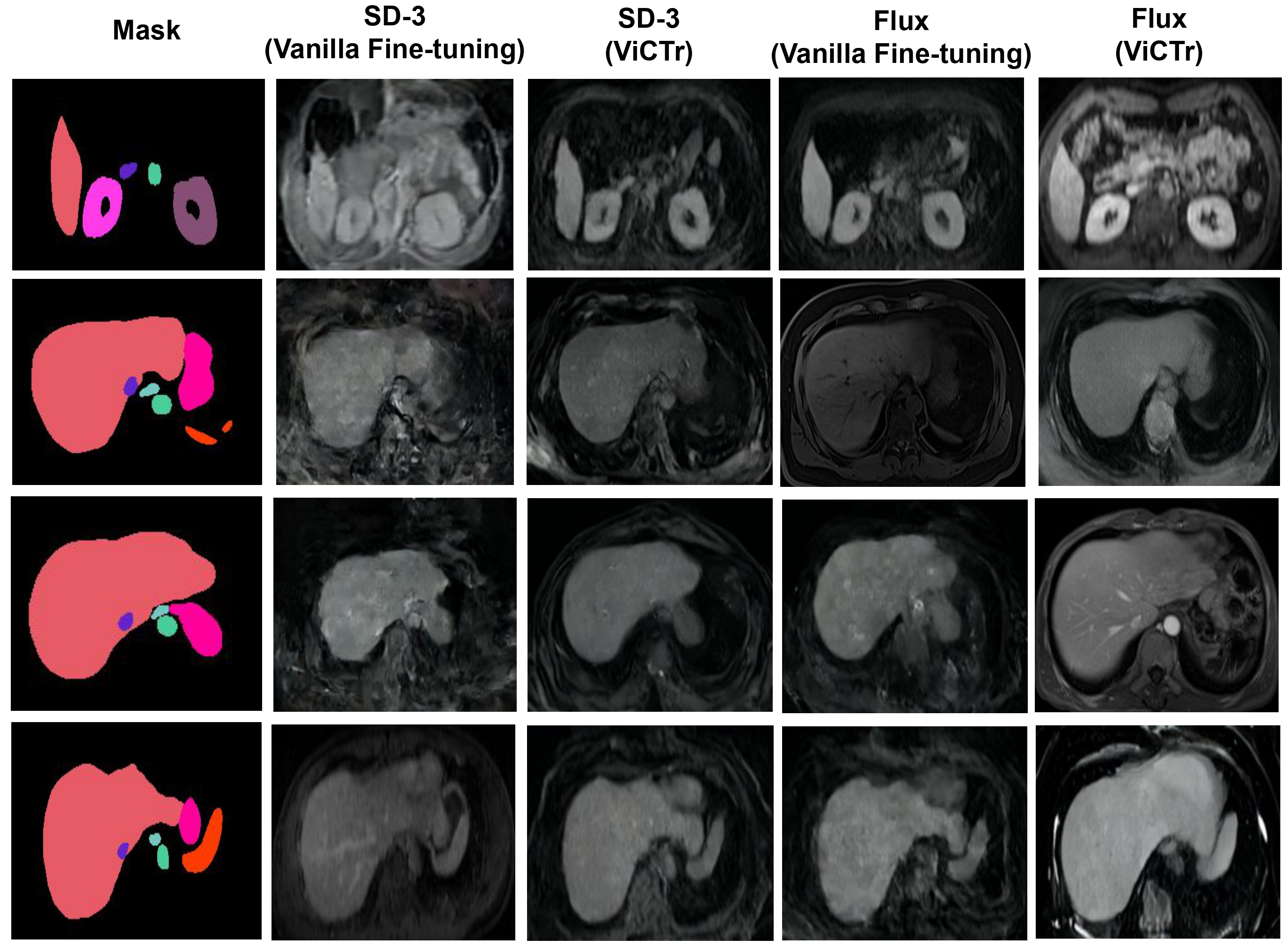}
        \caption{Diffusion-Based Image Generation on AMOS Dataset. From left to right: segmentation mask, SD-3 with Vanilla fine-tuning,  SD-3 with our method, FLUX with Vanilla fine-tuning, FLUX with our method.}
        \label{fig:AMOS}
\end{figure}

    \begin{table*}[]
        \centering
        \scriptsize
        \begin{tabular}{cllllllll}
        \hline
        \multirow{2}{*}{\textbf{Baselines}}                                    & \multicolumn{2}{c}{\textit{\textbf{BTCV Dataset (CT Generation)}}}                   & \multicolumn{2}{c}{\textit{\textbf{AMOS Dataset (MRI Generation)}}}          & \multicolumn{2}{c}{\textit{\textbf{CirrMRI600+ (Pathology Generation)}}}     & \multicolumn{2}{l}{\textbf{Diffusion Steps $|$ Inference Time}} \\
                                                                               & \multicolumn{1}{c}{Fine-tuned Vanilla} & \multicolumn{1}{c}{\textbf{ViCTr}}         & \multicolumn{1}{c}{Fine-tuned Vanilla} & \multicolumn{1}{c}{\textbf{ViCTr}} & \multicolumn{1}{c}{Fine-tuned Vanilla} & \multicolumn{1}{c}{\textbf{ViCTr}} & Fine-tuned Vanilla               & \textbf{ViCTr}                    \\
                                                                  \hline
        \textbf{Stable Diffusion \cite{Stable-Diffusion-1.5}} & 25.44 / 19.67                           & \textbf{21.98 / 19.02}                     & 25.43 / 21.76                           & \textbf{20.37 / 19.11}             & 28.34 / 23.43                           & \textbf{25.57 / 21.46}             & 40 $|$ 13.76                      & \textbf{4 $|$ 3.12}               \\
        \textbf{Stable Diffusion XL \cite{SD-XL}}             & 23.47 / 18.21                           & \textbf{20.33 / 17.44}                     & 24.11 / 20.23                           & \textbf{19.44 / 18.45}             & 27.34 / 22.11                           & \textbf{24.02 / 20.76}             & 30 $|$ 14.55                      & \textbf{4 $|$ 3.45}               \\
        \textbf{Stable Diffusion-3 \cite{Stable-Diffusion-3}} & 19.07 / 16.22                           & \textbf{17.37 / 16.02}                     & 22.32 / 19.76                           & \textbf{18.02 / 19.08}             & 24.49 / 21.78                           & \textbf{21.28 / 19.34}             & 50 $|$ 18.98                      & \textbf{3 $|$ 2.78}               \\
        \textbf{Pixart-alpha \cite{pixart-sigma}}             & 21.32 / 17.09                           & \textbf{19.22 / 16.96}                     & 23.78 / 20.04                           & \textbf{18.76 / 18.56}             & 26.06 / 20.07                           & \textbf{23.04 / 18.92}             & 25 $|$ 10.67                      & \textbf{3 $|$ 1.74}               \\
        \textbf{Flux \cite{fluxdev}}                          & 15.52 / 15.01                           & \multicolumn{1}{r}{\textbf{13.28 / 14.08}} & 19.02 / 18.28                           & \textbf{15.55 / 16.58}             & 22.46 / 18.88                           & \textbf{19.96 / 17.01}             & 30 $|$ 15.66                      & \textbf{3 $|$ 2.87}               \\ \hline
        \end{tabular}
        \caption{Quantitative results on CT, MRI, and Cirrhosis generation. All baselines were pre-trained on ATLAS-8k and fine-tuned on target datasets. Metrics reported are Fréchet Inception Distance (FID) and Medical FID (MFID), shown as FID/MFID, with inference time in seconds. Lower values indicate better performance.}

    \label{tab:generation_results}
    \end{table*}

    \begin{table*}[t]
    \centering
    \scriptsize
    \renewcommand{\arraystretch}{1.1}
    \begin{tabular}{ccccccccc}
    \toprule
    & \multicolumn{2}{c}{\textbf{Original Dataset (Org.)}} & \multicolumn{2}{c}{\textbf{Augmentation}} & \multicolumn{2}{c}{\textbf{Org. + Synth. by FLUX 30\% (vanilla Fine-tuning)}} & \multicolumn{2}{c}{\textbf{Org. + Synth. by FLUX 30\% (\textbf{ViCTr})}} \\
    \textbf{Baselines} & mDSC ($\uparrow$) & mHD95 ($\downarrow$) & mDSC ($\uparrow$) & mHD95 ($\downarrow$) & mDSC ($\uparrow$) & mHD95 ($\downarrow$) & mDSC ($\uparrow$) & mHD95 ($\downarrow$) \\
    \midrule
    \multicolumn{9}{c}{\cellcolor[HTML]{EFEFEF}\textit{BTCV dataset Segmentation results}} \\
    \midrule
    UNet \cite{ronneberger2015u} & 76.72 & 34.42 & 78.45 & 33.17 & 79.32 & 32.37 & 81.22 & 30.17 \\
    TransUnet \cite{chen2021transunet} & 85.52 & 32.33 & 87.01 & 31.43 & 87.54 & 30.11 & 89.78 & 29.12 \\
    nnUnet \cite{isensee2021nnu} & 80.48 & 30.19 & 82.54 & 31.02 & 83.37 & 29.01 & 85.19 & 27.77 \\
    nnFormer \cite{zhou2021nnformer} & 83.47 & 30.01 & 83.98 & 30.88 & 85.88 & 28.78 & 87.72 & 26.22 \\
    MedSegDiff \cite{wu2024medsegdiff} & 87.91 & 27.67 & \textcolor{orange}{88.65} & \textcolor{orange}{26.72} & \textcolor{darkgreen}{89.78} & \textcolor{darkgreen}{25.52} & \textcolor{blue}{91.92} & \textcolor{blue}{23.31} \\
    \midrule
    \multicolumn{9}{c}{\cellcolor[HTML]{EFEFEF}\textit{AMOS dataset Segmentation results}} \\
    \midrule
    UNet \cite{ronneberger2015u} & 68.92 & 34.57 & 70.55 & 32.32 & 71.02 & 31.78 & 73.34 & 29.11 \\
    TransUnet \cite{chen2021transunet} & 71.33 & 33.12 & 72.56 & 31.09 & 73.43 & 29.19 & 77.54 & 27.56 \\
    nnUnet \cite{isensee2021nnu} & 73.33 & 32.32 & 74.11 & 30.19 & 75.68 & 28.75 & 78.29 & 26.32 \\
    nnFormer \cite{zhou2021nnformer} & 75.78 & 31.19 & 76.44 & 29.76 & 77.77 & 27.57 & 81.32 & 24.41 \\
    MedSegDiff \cite{wu2024medsegdiff} & 76.83 & 29.92 & \textcolor{orange}{78.38} & \textcolor{orange}{26.99} & \textcolor{darkgreen}{79.03} & \textcolor{darkgreen}{25.45} & \textcolor{blue}{84.02} & \textcolor{blue}{22.18} \\
    \midrule
    \multicolumn{9}{c}{\cellcolor[HTML]{EFEFEF}\textit{CirrMRI600+ dataset Segmentation results}} \\
    \midrule
    UNet \cite{ronneberger2015u} & 68.74 & 36.73 & 69.38 & 35.43 & 70.12 & 35.11 & 73.39 & 32.11 \\
    TransUnet \cite{chen2021transunet} & 70.77 & 35.42 & 70.98 & 34.01 & 71.78 & 33.92 & 74.56 & 31.09 \\
    nnUnet \cite{isensee2021nnu} & 71.02 & 34.35 & 72.49 & 32.78 & 73.56 & 31.27 & 78.89 & 30.25 \\
    nnFormer \cite{zhou2021nnformer} & 74.88 & 33.78 & 75.23 & 31.88 & 76.45 & 31.09 & 79.44 & 29.78 \\
    MedSegDiff \cite{wu2024medsegdiff} & 76.92 & 30.79 & \textcolor{orange}{77.11} & \textcolor{orange}{30.34} & \textcolor{darkgreen}{78.03} & \textcolor{darkgreen}{29.89} & \textcolor{blue}{81.37} & \textcolor{blue}{27.34} \\
    \bottomrule
    \end{tabular}
    \caption{Segmentation results on BTCV, AMOS, and CirrMRI600+ using various training settings and baselines. Metrics reported are mDSC (\%) and mHD95 (mm). Top three results per setting are highlighted: \textcolor{blue}{blue} (1st), \textcolor{darkgreen}{green} (2nd), and \textcolor{orange}{orange} (3rd).}
    
    \label{tab:segmentation_results}
    \end{table*}

    \textbf{Synthetic Data Generation Results.} 
 Table \ref{tab:generation_results} summarizes our comparative evaluation of synthetic medical image generation under vanilla fine-tuning versus the proposed \textbf{ViCTr} pipeline, spanning multiple datasets (BTCV, AMOS, CirrMRI600+) and diffusion backbones (Stable Diffusion, SD-XL, SD-3, Pixart-alpha, Flux). While vanilla fine-tuning applies a direct end-to-end approach after ATLAS-8k pre-training, it generally yields higher Fréchet Inception Distance (FID) and Medical FID (MFID) values across all datasets. Conversely, \textbf{ViCTr }integrates domain-specific loss functions and specialized architectural settings, consistently achieving lower FID/MFID scores and producing more realistic medical images.

To address the well-known limitation of FID (which uses ImageNet features), we adopt M3D-CLIP \cite{bai2024m3d} to compute a more domain-specific MFID. As shown in the table, ViCTr provides notable gains: for example, Stable Diffusion under ViCTr achieves FID/MFID of 21.98/19.02 (BTCV), 20.37/19.11 (AMOS), and 25.57/21.46 (CirrMRI600+). These results represent substantial improvements over vanilla fine-tuning and demonstrate consistent performance boosts across all tested architectures. Such gains highlight the robustness of ViCTr’s two-stage framework in generating clinically meaningful synthetic images for both CT and MRI modalities.

    \textbf{Segmentation Results.} To assess the quality and utility of ViCTr-generated synthetic data, we performed segmentation experiments on three datasets—BTCV, AMOS, and CirrMRI600+—tracking mean Dice Similarity Coefficient (mDSC) and mean Hausdorff Distance 95 (mHD95). Higher mDSC and lower mHD95 respectively indicate improved overlap accuracy and spatial precision. We examined four training configurations: (1) baseline using original data only, (2) standard augmentation (Random Crop, Rotate, Blur, Affine, Geometric Distortion at 0.4 probability), (3) 30\% synthetic data via vanilla fine-tuning, and (4) 30\% synthetic data from \textbf{ViCTr}. Table \ref{tab:segmentation_results} highlights consistent performance gains from ViCTr across all datasets and backbone architectures. Notably, CirrMRI600+—a liver cirrhosis dataset—exhibited the most significant improvements, reflecting ViCTr’s capacity to capture complex pathological features. These results confirm that ViCTr-generated images significantly boost segmentation metrics (mDSC and mHD95), highlighting the practical value of our two-stage framework for medical data augmentation.

    \textbf{Efficiency of \textbf{ViCTr}.} Table \ref{tab:generation_results} summarizes the number of diffusion steps and average inference times (in seconds) for vanilla fine-tuning versus the \textbf{ViCTr} framework. Across all evaluated models, ViCTr consistently reduces the required steps, thereby shortening inference duration without compromising image fidelity. This efficiency stems from two key innovations: (1) Tweedy’s formula to approximate $Z_0$ and streamlines the reconstruction process, and (2) Two-Stage Training, enabling the model to learn both anatomical consistency and pathology-specific variations in a targeted manner. By minimizing extraneous steps, \textbf{ViCTr} can generate high-quality synthetic images more rapidly, making it a viable solution for resource-constrained clinical or research environments where computational overhead is a critical consideration.

\subsection{Qualitative Results}
\label{subsec:qualitative_results_Section}

We conducted extensive qualitative evaluations of \textbf{ViCTr} across multiple modalities (CT and MRI) and tasks, covering both anatomical and pathological image synthesis.

\textbf{Anatomical and Pathology-Driven Image Generation.} Unlike prior methods that primarily generate normal anatomical scans or isolated tumors, ViCTr extends synthesis capabilities to complex pathologies, such as liver cirrhosis. Figure \ref{fig:cirrosis} shows pairs of non-cirrhotic MRIs alongside their segmentation masks and the corresponding \textbf{ViCTr}-generated cirrhotic images. Notably, the synthesized images retain essential anatomical structures while introducing realistic cirrhotic texturing, an advancement that enables more diverse data augmentation and supports research into disease progression. These high-fidelity synthetic samples offer potential to enrich clinical training datasets, advance diagnostic algorithms, and support detailed analyses of pathology progression.
 
\begin{figure*}[]
    \centering
    \includegraphics[width=\textwidth]{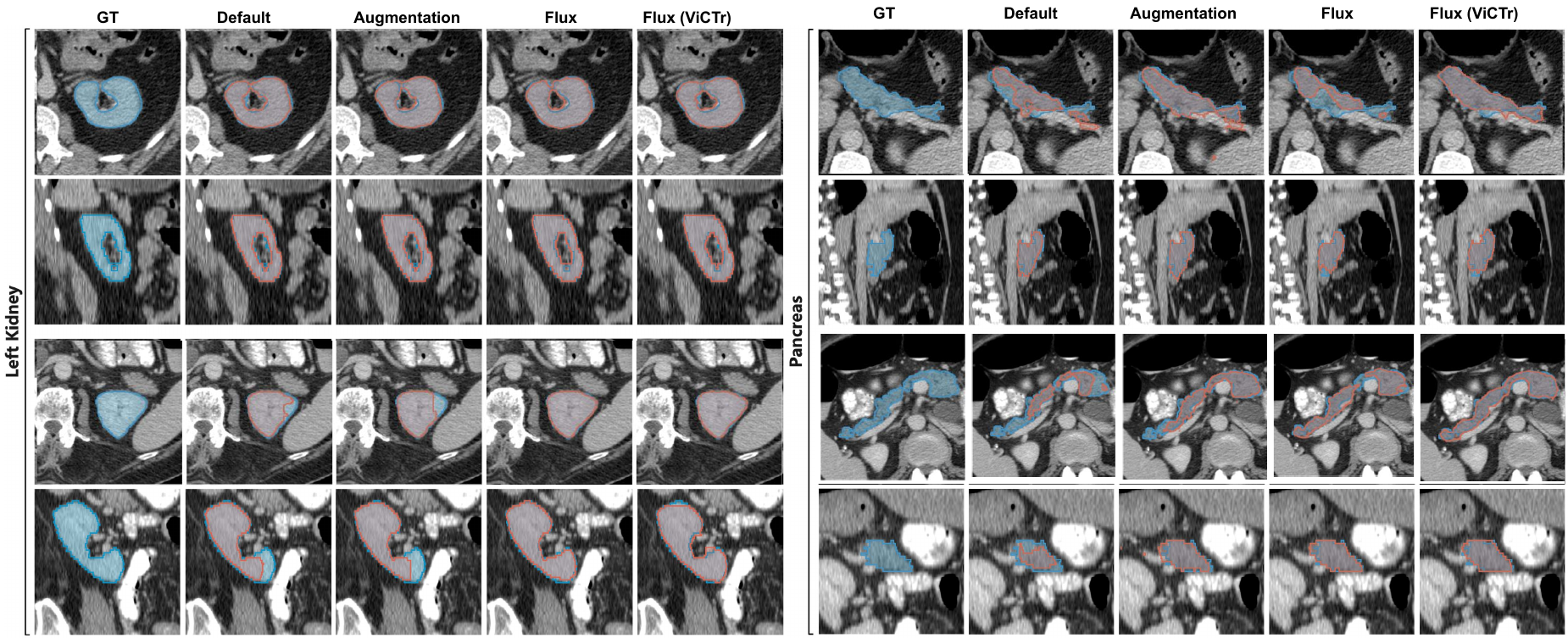}
        \caption{Segmentation results on BTCV and AMOS datasets' left kidney and pancreas. We compare the ground truth (blue), models trained on the default dataset, augmented dataset, default dataset plus 30\% synthetic data generated by FLUX with standard fine-tuning, and default dataset plus 30\% synthetic data generated by FLUX with \textbf{ViCTr} (red).}
    \label{fig:overview}

    \vspace{-4pt}
\end{figure*}

\textbf{Qualitative Segmentation Outcomes.} To further showcase \textbf{ViCTr}’s impact on downstream tasks, we visually assessed segmentation quality on BTCV and AMOS (Figure~\ref{fig:overview}). Focusing on regions like the left kidney and pancreas, models trained with \textbf{ViCTr}-augmented data show sharper boundaries and more precise delineation than standard fine-tuning, indicating improved spatial learning.

\textbf{Comparative Performance on BTCV and AMOS.} Figures \ref{fig:BTCV} (BTCV) and \ref{fig:AMOS} (AMOS) compare diffusion-based image generation results using Stable Diffusion v3 (SD-3) and Flux backbones. On BTCV—a dataset with limited samples and larger spatial dimensions—vanilla fine-tuning with SD-3 often yields poor mask adherence, suggesting difficulties in learning accurate spatial constraints. In contrast, the same model shows markedly improved mask alignment on the AMOS dataset, attributed to its larger training set and more manageable image dimensions.

Regardless of these dataset-specific variations, \textbf{ViCTr} achieves consistently high-quality outputs with robust mask adherence and anatomical accuracy across both BTCV and AMOS, underscoring its enhanced spatial learning and resilience. By maintaining performance even under data-sparse conditions, \textbf{ViCTr} demonstrates a clear advantage over conventional fine-tuning approaches, which can exhibit greater sensitivity to dataset size and complexity.

Second, we evaluate the impact of \textbf{ViCTr} on segmentation tasks, illustrated in Figure \ref{fig:overview}. Both BTCV and AMOS datasets encompass various abdominal structures (e.g., left kidney, pancreas), and our results highlight how \textbf{ViCTr}-based training yields notably more precise organ boundaries compared to alternative methods. This improvement is particularly evident in challenging regions where standard segmentation often struggle, highlighting \textbf{ViCTr}’s strength in enhancing spatial delineation and anatomical accuracy.

\subsection{Ablation Studies}
\label{sec:ablation_study_Section}

    \begin{table}[h]
        \centering
        \resizebox{\columnwidth}{!}{%
        \begin{tabular}{cccc}
        \hline
        \multicolumn{4}{c}{\textbf{Ablation Based on Pre-training of Model}}                                                                                                                \\ \hline
        \multicolumn{2}{c}{ViCTr (Without Stage-1)}                                                                              & \multicolumn{2}{c}{17.33}                                \\ \hline
        \multicolumn{4}{c}{\textbf{Ablation Based on Rectified Flow Algorithms}}                                                                                       \\ \hline
        \multicolumn{2}{c}{ViCTr (Without Proposed Tweedies Formula)}                                                            & \multicolumn{2}{c}{18.78}                                \\
        \multicolumn{2}{c}{ViCTr (With Reflow)}                                                                        & \multicolumn{2}{c}{20.19}                                \\
        \multicolumn{2}{c}{\begin{tabular}[c]{@{}c@{}}ViCTr (With Flow Straight and Fast)\end{tabular}}             & \multicolumn{2}{c}{21.37}                                \\
        \multicolumn{2}{c}{\begin{tabular}[c]{@{}c@{}}ViCTr (With Distribution Matching Distillation)\end{tabular}} & \multicolumn{2}{c}{22.33}                                \\ \hline
        \multicolumn{2}{c}{\textbf{Ablation Study Based on Loss Functions}}                                                      & \multicolumn{2}{c}{\textbf{Ablation Based on LoRA Rank}} \\ \hline
        $L_{diff}$                                                          & \multicolumn{1}{c|}{18.77}                         & r = 8                        & 18.46                     \\
        $L_{diff} + L_{spatial}$                                            & \multicolumn{1}{c|}{18.21}                         & r = 16                       & 17.52                     \\
        $L_{diff} + L_{consistancy}$                                        & \multicolumn{1}{c|}{17.02}                         & r = 32                       & 16.66                     \\
        $L_{diff} + L_{spatial} + L_{consistancy}$                          & \multicolumn{1}{c|}{15.55}                         & r = 64                       & 15.55                     \\ \hline
        \end{tabular}
        }
        \caption{Ablation study for the effect of model pretraining, loss functions, and LoRA rank settings are shown. 
        \label{tab:ablation}} 
    \end{table}

To rigorously validate our ViCTr framework, we conducted a series of ablation experiments on the BTCV dataset (Table \ref{tab:ablation}), focusing on four key aspects: pretraining, rectified flow algorithms, loss function components, and LoRA rank selection. We used FID (Fréchet Inception Distance) as our primary metric to gauge how each architectural choice affects both anatomical fidelity and pathological realism.

\textbf{Impact of Pretraining.} We first examined the necessity of Stage 1 pretraining by comparing our full, two-stage pipeline to a variant that proceeds directly to downstream fine-tuning. Omitting Stage 1 degrades the model’s FID score from 15.55 to 17.33 (Table \ref{tab:ablation}), underscoring the vital role of an anatomical prior. These findings affirm that establishing robust structural representations in Stage 1 is crucial for achieving high-quality medical image synthesis.

\textbf{Rectified Flow Algorithms.} We next assessed our rectified flow formulation, augmented with Tweedie’s formula, against alternative rectified flow–based models, including ReFlow, Flow Straight and Fast, and Distribution Matching Distillation. Removing Tweedie’s correction alone increased the FID to 18.78, highlighting its significance for trajectory alignment. Across all baselines, we observed FID deteriorations, ranging from 13.28 to as high as 22.33, thereby demonstrating that our proposed method provides tighter distribution matching and superior synthesis fidelity.

  %  \textbf{Ablation Based on Pretraining.}     Our first ablation experiment examines the necessity of Stage-1 pretraining by comparing our full framework against a variant that proceeds directly to downstream fine-tuning. When Stage-1 pretraining is removed, the model's FID score deteriorates to $17.33$ (Table~\ref{tab:ablation}), representing a substantial degradation in synthesis quality. This performance gap validates a key insight of our approach: establishing robust anatomical priors through pretraining is essential for high-quality medical image synthesis. The significant performance difference demonstrates that the two-stage design is not merely an implementation choice but a major requirement for achieving state-of-the-art synthesis quality.

  %  \textbf{Ablation Based on Rectified Flow Algorithms.} \textcolor{blue}{To assess the effectiveness of our Rectified Flow optimization with Tweedie’s formula, we compare it against other state-of-the-art rectified flow–based algorithms. As shown in Table \ref{tab:ablation}, omitting Tweedie’s formula leads to an increase in FID to 18.78, highlighting the importance of our approach in ensuring tighter trajectory alignment. For a fair comparison, we pretrain the ViCTr model using ReFlow, Flow Straight and Fast, and Distribution Matching Distillation methods, then fine-tune on the BTCV dataset. Across these baselines, we observe that the FID rises from 13.28 to as high as 22.33, underscoring the effectiveness of our proposed framework in closely aligning predicted and target distributions.}
    
    \textbf{Ablation Study Based on Loss Functions.} We then investigated each \textbf{ViCTr} loss component—diffusion loss, spatial consistency, and consistency loss—to determine their relative contributions: the baseline model, utilizing only diffusion loss, produced an FID of $18.77$ (Table~\ref{tab:ablation}). Incorporating spatial consistency loss yielded a moderate improvement to $18.21$, highlighting its role in maintaining structural integrity. The addition of consistency loss further enhanced performance, reducing the FID to $17.02$ by promoting image coherence across generations. When combining all three terms, we achieved an FID of 15.55, indicating that these components collectively offer significant improvements in anatomical accuracy and visual realism. %The optimal configuration emerged from the synergistic combination of all three losses, achieving a superior FID of $15.55$. This systematic analysis validates the essential contribution of each loss term to the overall generative performance, with their combined effect surpassing individual implementations.
    
    \textbf{Ablation Based on LoRA Rank.} Finally, we explored how varying the LoRA rank ($r$) influences generative performance. Starting at $r=8$ and incrementing upwards, we found $r=64$ to yield the best FID at $15.55$. This result suggests that higher-dimensional adaptation spaces enable more nuanced parameter updates during fine-tuning, thereby improving visual quality and fidelity. %This superior performance can be attributed to the enhanced expressiveness afforded by higher-dimensional adaptation spaces, enabling more nuanced feature extraction and transformation during the fine-tuning process. The results demonstrate that larger LoRA ranks facilitate more sophisticated parameter updates, leading to improved visual quality and fidelity in the generated outputs.
 
    \textbf{Visual Turing Tests.} To further validate clinical plausibility, three radiologists participated in Visual Turing Tests using 15 randomly generated MRI scans depicting varying levels of liver cirrhosis (mild, moderate, severe). All scans were uniformly judged to be clinically realistic, with identical outcomes observed when using additional random samples. Notably, the synthetic cirrhotic images correctly exhibited surface nodularity and textural irregularities, consistent with radiologic findings in mild-to-severe cirrhosis (Figure \ref{fig:cirrosis}). These results confirm ViCTr’s ability to synthesize pathology-specific features, enabling applications in training, algorithm development, and clinical research. \footnote{Code: \url{https://github.com/Onkarsus13/ViCTr-2D} \\ Weights: \url{https://huggingface.co/onkarsus13/ViCTr-2D}}
   % Three radiologists were asked to classify randomly generated 15 MRI scans depicting various levels of liver cirrhosis (mild, moderate, severe; in equal numbers) as either real or fake. The radiologists found all the scans to be realistic, with two participating in the Turing test by consensus and one conducting an independent assessment. The experiment was repeated using other random data sets, yielding similar results: all images were considered clinically feasible and real-like. In the synthetic cirrhotic images, the liver surface—typically smooth and lacking nodularity in healthy individuals—displays \textbf{surface nodularity} with varying intensities, accurately representing mild, moderate, and severe conditions . This surface nodularity is shown in Figure \ref{fig:cirrosis}, where the liver contour appears irregular. Moreover, changes in the \textbf{parenchymal texture} are evident, indicating an irregular or lobulated contour of the liver tissue. These changes correspond with radiological signs of cirrhosis, highlighting the model's ability to simulate clinically relevant pathological features.
\section{Conclusion}
\label{sec:conclusion_Section}
  We presented \textbf{ViCTr}, a novel two-stage framework designed to generate high-fidelity medical images by integrating robust anatomical pre-training with precise pathology-specific fine-tuning. By aligning Tweedie’s formula with linear projection methods used in flow matching, \textbf{ViCTr} maintains accurate initial distribution estimates even amid diffusion processes. This setup is further enhanced by LoRA adapters, which preserve essential anatomical information while flexibly adapting to diverse pathologies. Experimental results demonstrate that \textbf{ViCTr} consistently surpasses conventional fine-tuning strategies, reducing FID scores and producing clinically realistic outputs. Beyond data augmentation for segmentation and classification, \textbf{ViCTr} can be extended to tasks such as modality translation, contrast synthesis, and professional training, offering a powerful and versatile tool for advancing AI-driven medical imaging.

  \section*{Acknowledgments}
  This research is supported by the following NIH grants: R01-HL171376 and U01-CA268808.

{
    \small
    \bibliographystyle{ieeenat_fullname}
    \bibliography{main}
}

\maketitlesupplementary
\renewcommand{\thesection}{\Alph{section}}
\setcounter{section}{0}

\section{Quantitative}
\begin{figure}[b]
    \centering
    \includegraphics[width=1\linewidth]{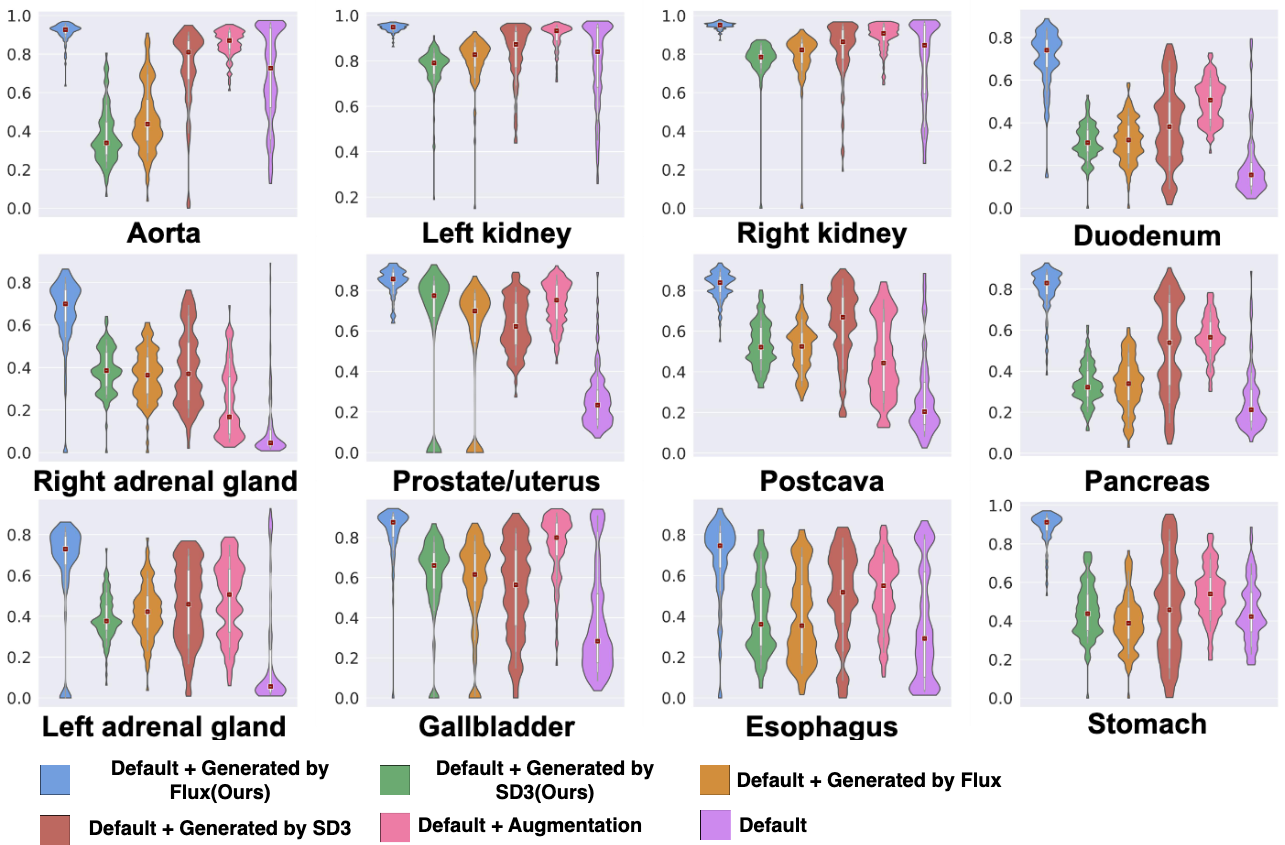}
    \caption{Segmentation Performance Comparison Using Violin Plots.}
    \label{fig:violingraph}
\end{figure}

Figure \ref{fig:violingraph} provides a quantitative comparison of segmentation models trained with various datasets, visualized using violin plots for organs such as the aorta, left kidney, right kidney, right adrenal gland, prostate, postcava, left adrenal gland, gallbladder, and esophagus. Models trained with our synthetic data generated by \textbf{ViCTr} show improved performance over those trained with default datasets, standard data augmentation, and synthetic data generated by standard fine-tuning methods. This further validates the efficacy of our approach in enhancing segmentation tasks.

\begin{figure}[]
        \centering
        \includegraphics[width=\columnwidth]{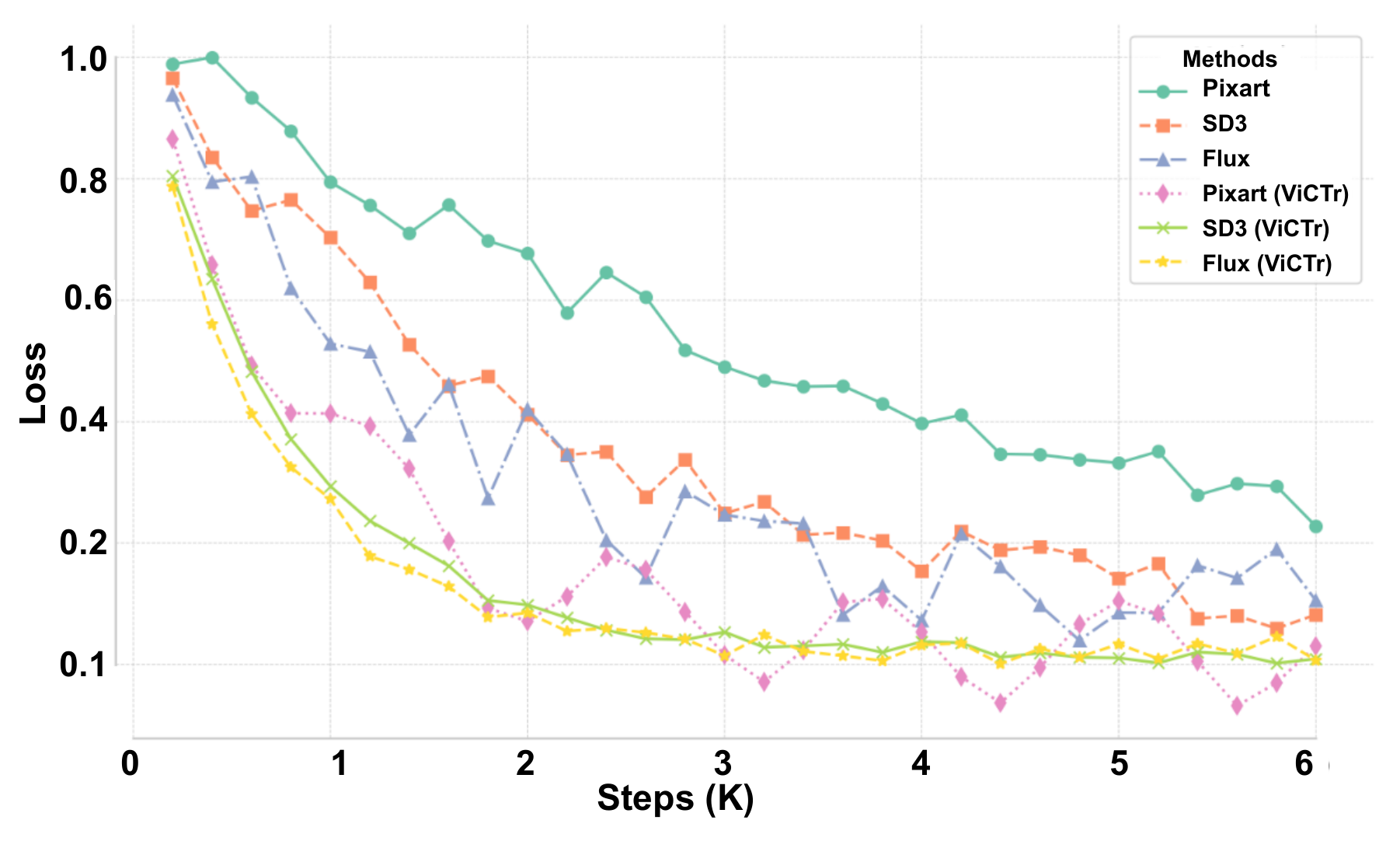}
        \caption{Convergence of models.}                \label{fig:convergence}
\end{figure}

\begin{figure}[t]
    \centering
    \includegraphics[width=1\linewidth]{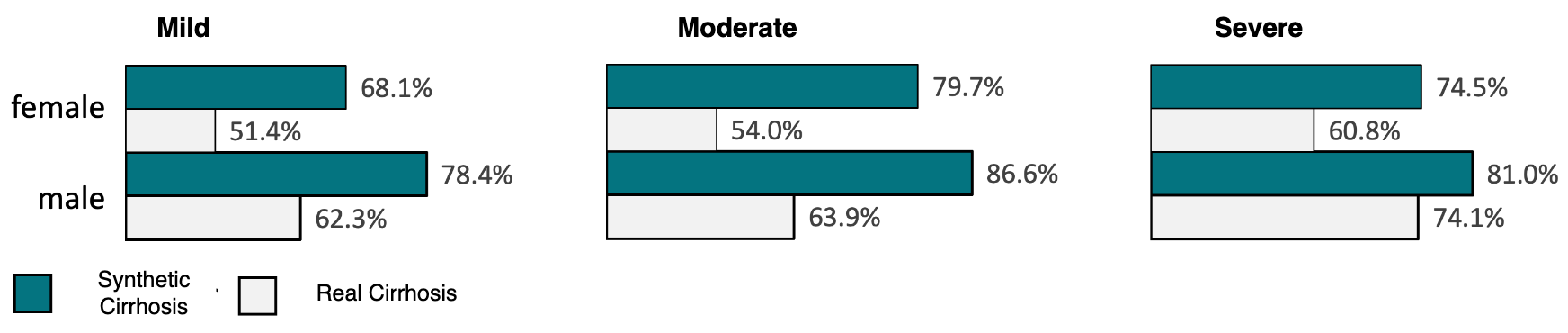}
    \caption{Comparison of severity levels (mild, moderate, severe) between real cirrhotic images and synthetic cirrhotic images generated by ViCTr for male and female subjects.}
    \label{fig:severity}
\end{figure}

Figure \ref{fig:severity} showcases the capability of \textbf{ViCTr} to control the severity of synthetic cirrhosis in generated images. We compare the severity levels mild, moderate, and severe between real cirrhotic images and our synthetic counterparts for both male and female subjects. The synthetic images accurately reflect the specified severity levels, and in some cases, they rank better in visual assessments than real images. This highlights the potential of our method for generating controlled pathological variations for training and diagnostic purposes.

%%%%%%%%%%%%%%%%%%%%%%%%%%%%%%%%%%%%%%%%%%%%%%%%%%%%%%%%%%%%%%%%%%%%%%%%%%%%%%%%%%%%%%

\textbf{Learning Efficiency of \textbf{ViCTr}-Enhanced Models:} Figure~\ref{fig:convergence} presents a comprehensive analysis of model convergence across 30 training steps, comparing \textbf{ViCTr} against baseline approaches. The results demonstrate \textbf{ViCTr's} superior convergence characteristics and learning efficiency across multiple state-of-the-art architectures (Pixart, SD3, and Flux—using standard vanilla fine-tuning). Lower loss values indicate better convergence, with a steeper decline in the early steps suggesting faster learning. The baseline models (Pixart, SD3, and Flux) trained with vanilla fine-tuning show a gradual decrease in loss but maintain relatively higher loss values throughout the training steps. For example, Pixart has the slowest convergence, with its loss remaining comparatively high even after 30 steps. In contrast, the \textbf{ViCTr}-enhanced models demonstrate much faster convergence rates and achieve significantly lower loss values. The consistent performance improvements across different architectures (Pixart, SD3, and Flux) further demonstrate the versatility and generalizability of our approach, establishing \textbf{ViCTr} as a powerful framework for advancing medical image synthesis. 
    
\textbf{Additional Segmentation Results:} 

\begin{table}[t]
\centering
\small
\setlength{\tabcolsep}{4pt}
\begin{tabular}{lcc}
\toprule
\textbf{Model} & \textbf{Vanilla FineTuning} & \textbf{ViCTr (Ours)} \\
\midrule
Stable Diffusion & 23.11 / 84.72 & 25.27 / 86.72 \\
Stable Diffusion-XL & 24.34 / 85.72 & 26.78 / 87.93 \\
Stable Diffusion-3 & 26.44 / 87.51 & 28.92 / 90.37 \\
Pixart & 26.32 / 88.21 & 31.09 / 91.33 \\
Flux & 27.51 / 90.21 & 33.33 / 94.05 \\
\bottomrule
\end{tabular}
\caption{PSNR / SSIM (\%) comparison between Vanilla FineTuning and our ViCTr across diffusion models on CirrMRI600+.}
\label{tab:psnr_ssim_iccv}
\end{table}

\begin{figure*}[t]
    \centering
    \includegraphics[width=1\linewidth]{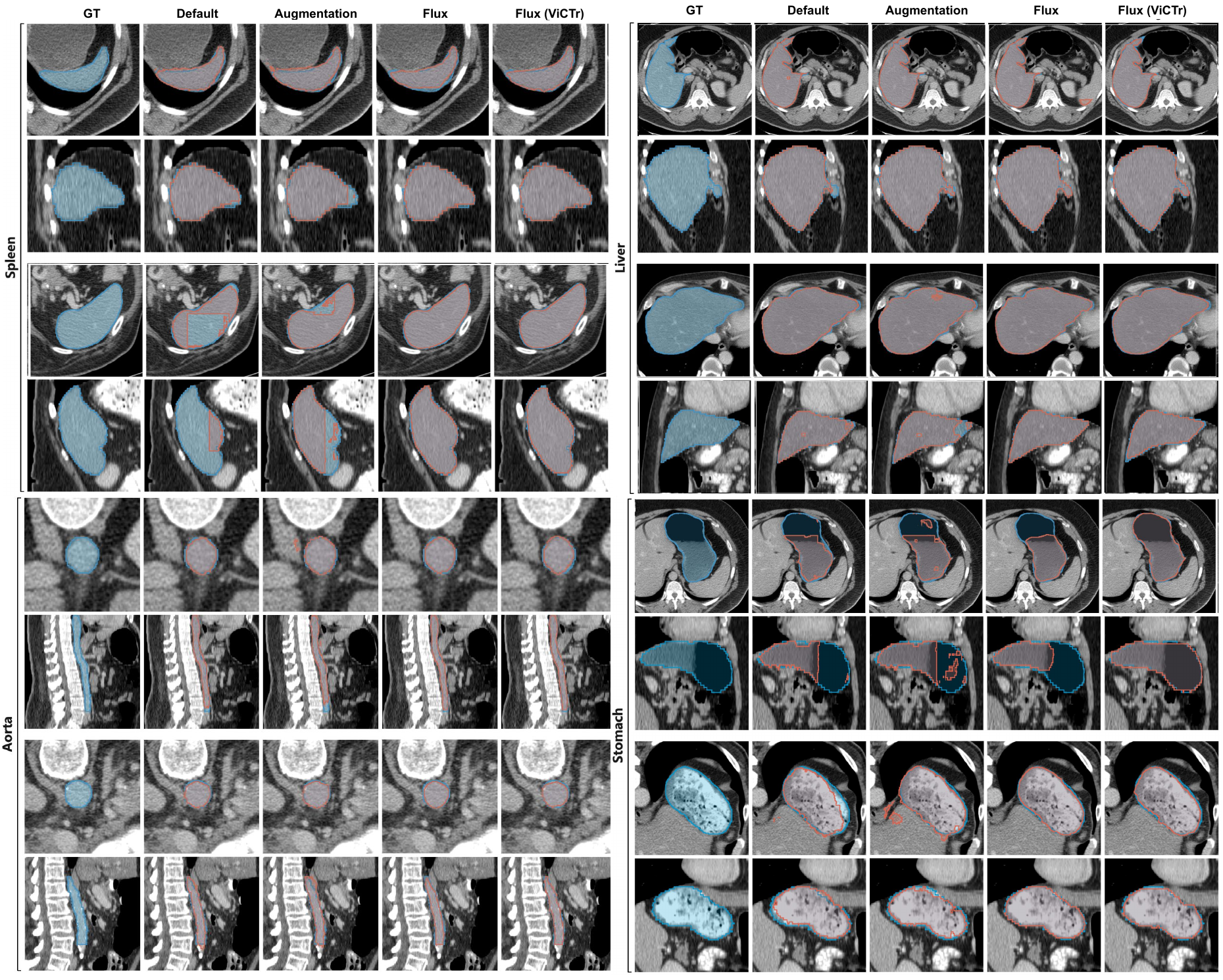}
    \caption{Segmentation results for comparison across various methods}
    \label{fig:segresults}
\end{figure*}

We present extended visual results showcasing segmentation performance on complex organs such as the spleen, liver, aorta, and stomach. As depicted in Figure \ref{fig:segresults}, our method, which leverages synthetic data generated via the Flux (\textbf{ViCtr}) framework, demonstrates superior alignment with ground truth (GT) segmentation. Notably, the quality and consistency of the predicted masks across all four organ classes are on par with GT annotations. These results highlight the efficacy of our approach in capturing intricate organ structures with high precision and robustness.

\subsection*{Modality Translation Results on CirrMRI600+}

\subsubsection*{Experimental Setup}
To evaluate cross-modality translation performance, see in Table~\ref{tab:psnr_ssim_iccv}, we conducted experiments using the paired T1–T2 volumes from the CirrMRI600+ dataset. The goal was to synthesize target modality (T2-weighted) images conditioned on anatomical features from the source modality (T1-weighted) using text-based prompts such as \textit{``Generate the pathology on T2-weighted MRI''}. 

We assessed both structural preservation and pathological fidelity of the translated outputs. Quantitative evaluation was carried out using Peak Signal-to-Noise Ratio (PSNR) and Structural Similarity Index Measure (SSIM), comparing synthesized T2 volumes against ground truth.

\subsubsection*{Results}
Our proposed ViCTr framework consistently outperformed baseline diffusion-based models across all metrics, demonstrating superior anatomical consistency and modality-specific detail reconstruction. These findings emphasize the potential of ViCTr in downstream clinical applications such as modality harmonization, synthetic augmentation, and diagnostic support.

\begin{table}
\begin{tabular}{cc}
\hline
\textbf{Training H-Prameters} & \textbf{Values} \\ \hline
Learning Rate                 & 1.00E-04        \\
Gradient Accumalation Steps   & 8               \\
Batch Size Per GPU            & 2               \\
Optimizer                     & AdamW           \\
Lr-Schedular                  & Cosine          \\
Epochs                        & 40              \\
Noise Schedular               & FlowMatching    \\
Diffusion Steps               & 100             \\
Training Precision            & BFloat16        \\
GPUs                          & 8 x 8 A100      \\
Text Encoders                 & T5-XXXL         \\
Time Embedding Size           & 512             \\
Gradient Clipping             & 2.5             \\
Max Text Length               & 200             \\
Embedding Size                & 4096            \\
CFG Scale                     & 10.5            \\
Positional Encodings          & RoPE            \\ \hline
\end{tabular}
\label{tab:h_parameters}
\caption{Hyper-parameters used to train models}
\end{table}

\section{Qualitative}
\begin{figure*}[]
    \centering
    \includegraphics[width=1\linewidth]{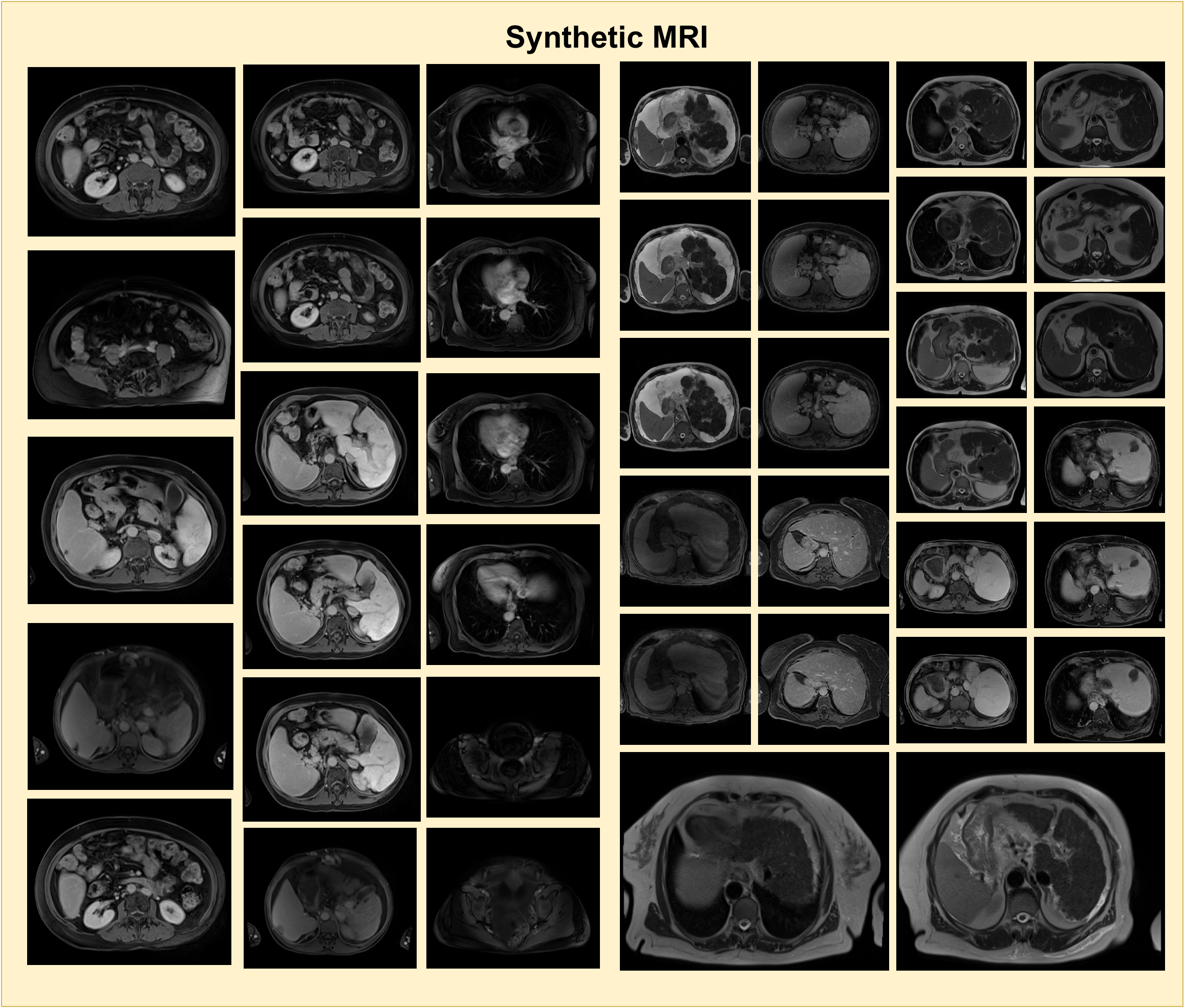}
    \caption{Synthetically generated MRI images using  \textbf{ViCTr}.}
    \label{fig:synth_imgs}
\end{figure*}

This Figure \ref{fig:synth_imgs} presents a additional visual results of synthetic MRI images generated using \textbf{ViCTr}.

\section{Training and Implementation Details} \label{sec:training_details}

\textbf{Pre-training.} We pre-trained \textbf{ViCTr} Stage-1 using a rectified flow strategy, with the maximum diffusion steps set to 100. The Atlas-8K dataset was used as the foundational dataset, and training was performed at an image resolution of $256 \times 256$. We employed a batch size of $8$, with gradient accumulation over 8 steps. Optimization was carried out using the Adam optimizer with an initial learning rate of $1 \times 10^{-5}$, managed by a cosine annealing scheduler to ensure a smooth decay of the learning rate over time. The pre-training phase was conducted on 4 nodes, each equipped with 8 Nvidia A100 GPUs 80GB each, and completed in approximately 52 hours.

\textbf{Fine-tuning.} For fine-tuning, we initialized \textbf{ViCTr} Stage-2 with the pre-trained weights from Stage-1 and configured it for the downstream tasks of CT, MRI, and pathological image generation. Fine-tuning was carried out at a $256 \times 256$ resolution, using a batch size of 4 with gradient accumulation over 12 steps. The Adam optimizer was used but with a higher initial learning rate of $1 \times 10^{-4}$, and a cosine learning rate scheduler for adaptive adjustment throughout training. Fine-tuning was conducted on a 2-node setup, each equipped with 8 Nvidia A100 GPUs 80GB each. Given Table below shows  

\end{document}